arXiv:1610.06949v1 [stat.ML] 21 Oct 2016

# Mean-Field Variational Inference for Gradient Matching with Gaussian Processes

**Nico S. Gorbach*** NGORBACH@INF.ETHZ.CH

**Stefan Bauer*** BAUERS@INF.ETHZ.CH

**Joachim M. Buhmann** JBUHMANN@INF.ETHZ.CH

*Department of Computer Science, ETH Zurich*
*8092 Zurich, Switzerland*

## Abstract

Gradient matching with Gaussian processes is a promising tool for learning parameters of ordinary differential equations (ODE's). The essence of gradient matching is to model the prior over state variables as a Gaussian process which implies that the joint distribution given the ODE's and GP kernels is also Gaussian distributed. The state-derivatives are integrated out analytically since they are modelled as latent variables. However, the state variables themselves are also latent variables because they are contaminated by noise. Previous work sampled the state variables since integrating them out is *not* analytically tractable. In this paper we use mean-field approximation to establish tight variational lower bounds that decouple state variables and are therefore, in contrast to the integral over state variables, analytically tractable and even concave for a restricted family of ODE's, including nonlinear and periodic ODE's. Such variational lower bounds facilitate "hill climbing" to determine the maximum a posteriori estimate of ODE parameters. An additional advantage of our approach over sampling methods is the determination of a proxy to the intractable posterior distribution over state variables given observations and the ODE's.

## 1. Introduction

Parameter estimation for ordinary differential equations (ODE) is challenging due to the high computational cost of numerical integration. Different approaches based on minimizing the difference between the interpolated slopes and the time derivatives of the state variables in the ODE's go back to spline based methods (Varah (1982); Ramsay et al. (2007) or Campbell and Steele (2012)). However, these methods depend critically on additional regularization and selected initial values. In recent years approaches based on Gaussian process regression have been in the spotlight of the machine learning community (e.g. Calderhead et al., 2008; Niu et al., 2016; Wang and Barber, 2014; Dondelinger et al., 2013). An overview of the different approaches with a focus on the application for systems biology is provided in Macdonald and Husmeier (2015). These approaches enable Bayesian parameter inference without explicitly solving the ODE. However, in this previous work latent state variables were sampled, while we use mean-field approximations to establish tight variational lower bounds, that decouple the state variables and are therefore analytically tractable. Moreover we are able to determine a proxy to the intractable posterior distribution over state variables given the observations and the parameters of the ODE's.

*. The first two authors contributed equally to this work

We start by introducing the gradient matching model in section 2 and show in section 3 that integrating out the latent variables is analytically tractable for the state derivatives given the states but is intractable for both the state variables and states derivatives. In section 4 we describe the mechanism behind variational inference. In the subsequent section 5 we use mean-field variational inference to establish tight variational lower bounds that are analytically tractable since they decouple state variables. Experiments on simulated and real-world data in section 6 show the applicability of our results and we conclude by a discussion in section 7.

## 2. Gradient Matching with Gaussian Processes

In the following we use the description of Gradient Matching with Gaussian processes from Calderhead et al. (2008), as described in Dondelinger et al. (2013). A dynamical system is represented by a set of $K$ ordinary differential equations (ODE's) with model parameters $\boldsymbol{\theta}$ that describe the evolution of $K$ states $\mathbf{x}(t) = [x_1(t), x_2(t), \dots, x_K(t)]^T$ such that:

$$\dot{\mathbf{x}}(t) = \frac{d\mathbf{x}(t)}{dt} = \mathbf{f}(\mathbf{x}(t), \boldsymbol{\theta}). \tag{1}$$

In the following we consider only models of the form 1 with reactions based on mass-action kinetics. In particular we consider for each dimension only functions of the form

$$f(\mathbf{x}(t), \boldsymbol{\theta}) = \sum_{i=1}^{M} \theta_i \prod_{j \in \mathcal{S}} x_j \tag{2}$$

with $\mathcal{S} \subseteq \{1, \dots, K\}$ and $M$ describing the number of terms in each equation. This formulation includes models which exhibit periodicity as well as high non-linearities and especially physically realistic reactions in systems biology (Schillings et al., 2015).

A sequence of observations, $\mathbf{y}(t)$ are usually contaminated by some measurement error which we assume to be normally distributed with zero mean and variance for each of the $K$ states, i.e. $\mathbf{E} \sim \mathcal{N}(\mathbf{0}, \mathbf{D})$, with $\mathbf{D}_{ik} = \sigma_k^2 \delta_{ik}$. Thus for $N$ distinct time points the overall system may be summarized as:

$$\mathbf{Y} = \mathbf{X} + \mathbf{E},$$

where

$$\mathbf{X} = [\mathbf{x}(t_1), \dots, \mathbf{x}(t_N)] = [\mathbf{x}_1, \dots, \mathbf{x}_K]^T$$
$$\mathbf{Y} = [\mathbf{y}(t_1), \dots, \mathbf{y}(t_N)] = [\mathbf{y}_1, \dots, \mathbf{y}_K]^T,$$

where $\mathbf{x}_k = [x_k(t_1), \dots, x_k(t_N)]^T$ is the k'th state sequence and $\mathbf{y}_k = [y_k(t_1), \dots, y_k(t_N)]^T$ are the observations. Assuming a Gaussian process prior on state variables such that:

$$p(\mathbf{X} \mid \boldsymbol{\phi}) := \prod_k \mathcal{N}(\mathbf{0}, \mathbf{C}_{\boldsymbol{\phi}_k}) \tag{3}$$

where $\mathbf{C}_{\boldsymbol{\phi}_k}$ is a covariance matrix defined by some kernel with hyper-parameters $\boldsymbol{\phi}_k$, the $k$-th element of $\boldsymbol{\phi}$, we obtain a posterior distribution over state-variables:

$$p(\mathbf{X} \mid \mathbf{Y}, \boldsymbol{\phi}, \boldsymbol{\sigma}) = \prod_k \mathcal{N}\left(\boldsymbol{\mu}_k(\mathbf{y}_k), \boldsymbol{\Sigma}_k\right), \tag{4}$$

where $\boldsymbol{\mu}_k(\mathbf{y}_k) := \mathbf{C}_{\phi_k}(\mathbf{C}_{\phi_k} + \sigma_k^2\mathbf{I})^{-1}\mathbf{y}_k$ and $\boldsymbol{\Sigma}_k := \sigma_k^2\mathbf{C}_{\phi_k}(\mathbf{C}_{\phi_k} + \sigma_k^2\mathbf{I})^{-1}$. Notice that the posterior distribution over states in 4 conditions only on the observations $\mathbf{y}_k$ and not on the ODE's. The posterior distribution over states conditioned on the observations and the ODE's is even more desirable and will be approximated later in section 5.

We wish to determine the maximum a posteriori (MAP) estimate for the ODE parameters $\boldsymbol{\theta}$ given the observation $\mathbf{Y}$. The essence of using Gaussian processes to determine ODE parameters $\boldsymbol{\theta}$ is that we obtain a jointly Gaussian distribution over state variables $\mathbf{x}_k$ and their derivatives $\dot{\mathbf{x}}_k$ which is a consequence of the closure property of Gaussian distributions under differentiation. This allows us to obtain a posterior distribution over $\boldsymbol{\theta}$ given the ODE's (equation 1), the smoothed state and state derivatives. More precisely, the conditional distribution over state derivatives is:

$$p(\dot{\mathbf{X}} \mid \mathbf{X}, \boldsymbol{\phi}) = \prod_k \mathcal{N}(\dot{\mathbf{x}}_k \mid \mathbf{m}_k, \mathbf{A}_k), \tag{5}$$

where the mean and covariance is given by:

$$\begin{aligned} \mathbf{m}_k &:= {}'\mathbf{C}_{\phi_k}\mathbf{C}_{\phi_k}^{-1}\mathbf{x}_k \\ \mathbf{A}_k &:= \mathbf{C}''_{\phi_k} - {}'\mathbf{C}_{\phi_k}\mathbf{C}_{\phi_k}^{-1}\mathbf{C}'_{\phi_k}, \end{aligned}$$

$\mathbf{C}''_{\phi_k}$ denotes the auto-covariance for each state-derivative with $\mathbf{C}'_{\phi_k}$ and ${}'\mathbf{C}_{\phi_k}$ denoting the cross-covariances between the state and its derivative.

Assuming additive, normally distributed noise with state-specific error variance $\gamma_k$ in 1, we have:

$$p(\dot{\mathbf{X}} \mid \mathbf{X}, \boldsymbol{\theta}, \boldsymbol{\gamma}) = \prod_k \mathcal{N}\left(\dot{\mathbf{x}}_k \mid \mathbf{f}_k(\mathbf{X}, \boldsymbol{\theta}), \gamma_k\mathbf{I}\right). \tag{6}$$

A product of experts approach combines the ODE informed distribution of state-derivatives (distribution 5) with the smoothed distribution of state-derivatives (distribution 6):

$$p(\dot{\mathbf{X}} \mid \mathbf{X}, \boldsymbol{\theta}, \boldsymbol{\phi}, \boldsymbol{\gamma}) \propto p(\dot{\mathbf{X}} \mid \mathbf{X}, \boldsymbol{\phi})p(\dot{\mathbf{X}} \mid \mathbf{X}, \boldsymbol{\theta}, \boldsymbol{\gamma})$$

Calderhead et al. (2008) obtain the posterior distribution over ODE parameters by integrating over the state-derivatives:

$$p(\boldsymbol{\theta} \mid \mathbf{X}, \boldsymbol{\phi}, \boldsymbol{\gamma}) \propto p(\boldsymbol{\theta}) \int p(\dot{\mathbf{X}} \mid \mathbf{X}, \boldsymbol{\phi})p(\dot{\mathbf{X}} \mid \mathbf{X}, \boldsymbol{\theta}, \boldsymbol{\gamma})d\dot{\mathbf{X}},$$

which given the covariance matrix $\mathbf{C}_{\boldsymbol{\phi}}$, is analytically tractable and yields:

$$p(\boldsymbol{\theta} \mid \mathbf{X}, \boldsymbol{\phi}, \boldsymbol{\gamma}) \propto p(\boldsymbol{\theta}) \prod_k \mathcal{N}(\mathbf{f}_k(\mathbf{X}, \boldsymbol{\theta}) \mid \mathbf{m}_k, \boldsymbol{\Lambda}_k^{-1}), \tag{7}$$

where $\boldsymbol{\Lambda}_k^{-1} := \mathbf{A}_k + \mathbf{I}\gamma_k$ and $\mathbf{m}_k$ is defined as before.

Notice that distribution 7 conditions on the latent state variables $\mathbf{X}$. Calderhead et al. (2008) and Dondelinger et al. (2013) sample the latent variable from distribution 4 and subsequently sample the ODE parameters $\boldsymbol{\theta}$ according to distribution 7:

1. $\mathbf{X} \sim p(\mathbf{X} \mid \mathbf{Y}, \boldsymbol{\sigma}, \boldsymbol{\phi})$

2. $\boldsymbol{\theta}, \boldsymbol{\gamma} \sim p(\boldsymbol{\theta}, \boldsymbol{\gamma} \mid \mathbf{X}, \boldsymbol{\phi}, \boldsymbol{\sigma})$,

which is not directly feasible and requires a two-step sampling procedure.

Instead of sampling the state variables $\mathbf{X}$, our contribution is to integrate them out:

$$p(\boldsymbol{\theta} \mid \mathbf{Y}, \boldsymbol{\phi}, \boldsymbol{\gamma}) = \int p(\boldsymbol{\theta} \mid \mathbf{X}, \boldsymbol{\theta}, \boldsymbol{\phi}, \boldsymbol{\gamma}) p(\mathbf{X} \mid \mathbf{Y}, \boldsymbol{\phi}) d\mathbf{X} \tag{8}$$

which is in general *not* analytically tractable for non-linear ODE's $\mathbf{f}(\mathbf{X}, \boldsymbol{\theta})$ as in 1.

## 3. Couplings within and across States and State Derivatives

To summarize gradient matching methods based on Gaussian processes and highlight their differences, we consider the couplings in the joint posterior distribution over all unknowns, namely ODE parameters, states and their derivatives i.e.: $p(\boldsymbol{\theta}, \dot{\mathbf{X}}, \mathbf{X} \mid \mathbf{Y}, \boldsymbol{\phi}, \boldsymbol{\gamma})$.

1. **Time**: Same states e.g. $x_k$ are coupled between different time points i.e. $x_k(j) \not\perp\!\!\!\perp x_k(t), j \neq t$, which is due to the multivariate distribution over time points in the prior in 3.

2. **States**: In non-linear ODE's the states appear non-linearly in the joint posterior distribution and are thus coupled to each other i.e. $x_u(t) \not\perp\!\!\!\perp x_k(t), u \neq k$, despite the independence assumption of the prior in 3 and the likelihood. The same coupling applies to the occurrences of state derivatives at the same time point i.e. $\dot{x}_u(t) \not\perp\!\!\!\perp \dot{x}_k(t), u \neq k$, since the state derivatives $\dot{\mathbf{X}}$ in the joint distribution induce a coupling between the states, which persists even after integrating out the state derivatives!

3. **Derivatives**: The analytical integration of state derivatives is possible because their occurrences at the same time point are independent given the states i.e. $\dot{x}_u(t) \perp\!\!\!\perp \dot{x}_k(t) \mid \mathbf{X}, u \neq t$ as shown in 6. In fact the analogue independence applies to the states (i.e. $x_u(t) \perp\!\!\!\perp x_k(t) \mid \dot{\mathbf{X}}, u \neq t$). Thus we can integrate out the states analytically given the state derivatives and vice versa but one cannot integrate out both the state and state derivatives analytically for non-linear ODE's[1].

While $p(\boldsymbol{\theta}, \dot{\mathbf{X}}, \mathbf{X} \mid \mathbf{Y}, \boldsymbol{\phi}, \boldsymbol{\gamma})$ is an intractable distribution over all latent variables, the posterior over $\boldsymbol{\theta}$ is the one of highest interest. Wang and Barber (2014) integrate out the states and are able to do so analytically because they do not integrate over the joint posterior distribution. Consequently, after integrating out the states, Wang and Barber (2014) reintroduce them when conditioning on the ODE's which has lead to the controversy in the mechanistic modelling with Gaussian processes (Macdonald et al., 2015). Calderhead et al. (2008) analytically integrate out $\dot{\mathbf{X}}$, sample $\mathbf{X} \sim p(\mathbf{X} \mid \mathbf{Y}, \boldsymbol{\phi}, \boldsymbol{\sigma})$ and afterwards the parameters given the state variables, $\boldsymbol{\theta}, \boldsymbol{\gamma} \sim p(\boldsymbol{\theta}, \boldsymbol{\gamma} \mid \mathbf{X}, \boldsymbol{\phi}, \boldsymbol{\sigma})$. In this setup, sampling $\mathbf{X}$ is independent of $\boldsymbol{\theta}$, which implies that $\boldsymbol{\theta}$ and $\boldsymbol{\gamma}$, have no influence on the inference of the state variables. The desired feedback loop was closed by Dondelinger et al. (2013), integrating out the state derivatives and sampling from the joint posterior of $p(\boldsymbol{\theta} \mid \mathbf{X}, \boldsymbol{\phi}, \boldsymbol{\sigma})$. Our approach is motivated by the fact that the parameters of the ODE in equation 1 are the latent variables of highest interest. Thus we propose to simultaneously integrate out the state variables

1. For 1st order ODE's that are a linear combination of the states, the states appear linearly in the joint posterior distribution, in which case we can integrate them out analytically.

and state derivatives, which is enabled by the design of our proxy distribution 11 and mean-field variational inference in chapter 5. An additional advantage of our approach is that, in contrast to previous approaches, we are thus able to calculate a proxy distribution for the intractable posterior $p(\mathbf{X} \mid \mathbf{Y}, \boldsymbol{\theta}, \boldsymbol{\phi}, \gamma)$ in lemma 4.

## 4. Variational Inference

Variational inference transforms complex inference problems into optimization problems and tends to be faster than classical approaches based on sampling (Jordan et al., 1999; Opper and Saad, 2001). In recent years variational inference have been applied with great success e.g. in natural language processing , where it was used to analyze very large sets of documents like 1.8M *New York Times* or 3.8M *wikipedia* articles (Hoffman et al., 2013). In the following, we will use variational inference to infer the parameters of the dynamic systems described in equation 1, providing a fully Bayesian framework, even for distributions from which it is impossible to sample from.

The maximum a posteriori estimate (MAP) is given by:

$$\boldsymbol{\theta}^* := \arg\max_{\boldsymbol{\theta}} \ln \int p(\boldsymbol{\theta} \mid \mathbf{X}, \boldsymbol{\phi}, \gamma) p(\mathbf{X} \mid \mathbf{Y}, \boldsymbol{\phi}) d\mathbf{X}, \tag{9}$$

Our strategy is to establish variational lower bounds that decouple the distributions in the integral:

$$\begin{aligned}
&\ln \int p(\boldsymbol{\theta} \mid \mathbf{X}, \boldsymbol{\phi}, \gamma) p(\mathbf{X} \mid \mathbf{Y}, \boldsymbol{\phi}) d\mathbf{X} \\
&\overset{(a)}{=} - \int Q(\mathbf{X}) d\mathbf{X} \; \ln \frac{\int Q(\mathbf{X}) d\mathbf{X}}{\int p(\boldsymbol{\theta} \mid \mathbf{X}, \boldsymbol{\phi}, \gamma) p(\mathbf{X} \mid \mathbf{Y}, \boldsymbol{\phi}) d\mathbf{X}} \\
&\overset{(b)}{\geq} - \int Q(\mathbf{X}) \ln \frac{Q(\mathbf{X})}{p(\boldsymbol{\theta} \mid \mathbf{X}, \boldsymbol{\phi}, \gamma) p(\mathbf{X} \mid \mathbf{Y}, \boldsymbol{\phi})} d\mathbf{X} \\
&= H(Q) + \mathbb{E}_Q \ln p(\boldsymbol{\theta} \mid \mathbf{X}, \boldsymbol{\phi}, \gamma) + \mathbb{E}_Q \ln p(\mathbf{X} \mid \mathbf{Y}, \boldsymbol{\phi}) \\
&=: \mathcal{L}_Q(\boldsymbol{\theta})
\end{aligned} \tag{10}$$

where $H(Q)$ is the entropy. In (a) we introduce the auxiliary distribution $Q(\mathbf{X})$, $\int Q(\mathbf{X}) d\mathbf{X} = 1$ and in (b) we establish a lower bound using the log-sum inequality (i.e. $(\sum_i a_i) \log \frac{\sum_i a_i}{\sum_i b_i} \leq \sum_i a_i \log \frac{a_i}{b_i}$). Notice that the lower bound holds with equality whenever:

$$\begin{aligned}
Q^*(\mathbf{X}) &:= \frac{p(\boldsymbol{\theta} \mid \mathbf{X}, \boldsymbol{\phi}, \gamma) p(\mathbf{X} \mid \mathbf{Y}, \boldsymbol{\phi})}{\int p(\boldsymbol{\theta} \mid \mathbf{X}, \boldsymbol{\phi}, \gamma) p(\mathbf{X} \mid \mathbf{Y}, \boldsymbol{\phi}) d\mathbf{X}} \\
&\overset{(c)}{=} p(\mathbf{X} \mid \mathbf{Y}, \boldsymbol{\theta}, \boldsymbol{\phi}, \gamma),
\end{aligned}$$

where in (c) we use Bayes' rule. However $Q^*$ is analytically intractable because its normalization given by the integral 8 is in general analytically intractable. We therefore use mean-field approximation to determine a proxy distribution for $Q^*$ in section 5.

## 5. Mean-field Variational Inference

The aim of mean-field variational inference is to establish variational lower bounds that are analytically tractable by decoupling state variables across their time points. The decoupling of state variables is induced by designing a proxy[2] distribution $Q(\mathbf{X})$ which is restricted to the family of factorial distributions:

$$\mathcal{Q} := \left\{ Q : Q(\mathbf{X}) = \prod_k \prod_t q_{\boldsymbol{\psi}_{kt}}(x_k(t)) \right\}, \tag{11}$$

where $\boldsymbol{\psi}_{kt}$ are the variational parameters.

Note that the logarithm of integral 8 has the following identity:

$$\begin{aligned} &\ln \int p(\boldsymbol{\theta} \mid \mathbf{X}, \boldsymbol{\phi}, \boldsymbol{\gamma}) p(\mathbf{X} \mid \mathbf{Y}, \boldsymbol{\phi}) d\mathbf{X} \\ &= \mathcal{L}_Q(\boldsymbol{\theta}) + D_{KL}\left[Q(\mathbf{X}) \middle| \middle| p(\mathbf{X} \mid \mathbf{Y}, \boldsymbol{\theta}, \boldsymbol{\phi}, \boldsymbol{\gamma})\right] \end{aligned} \tag{12}$$

Since the Kullback-Leibler divergence $D_{KL}$ is always positive maximizing the lower bound $\mathcal{L}_Q(\boldsymbol{\theta})$ implicitly minimizes the Kullback-Leibler divergence in the identity 12.

To find the optimal factorial distribution, we maximize the lower bound 10 w.r.t. $q_{\boldsymbol{\psi}_{u\alpha}}(x_u(\alpha))$ observing the normalization constraint $\int q_{\boldsymbol{\psi}_{u\alpha}}(x_u(\alpha))\, dx_u(\alpha) = 1$:

$$\begin{aligned} 0 &\stackrel{!}{=} \frac{d}{dq_{\boldsymbol{\psi}_{u\alpha}}(x_u(\alpha))} \mathcal{L}_Q(\boldsymbol{\theta}) \\ &\quad + \lambda \left( \int q_{\boldsymbol{\psi}_{u\alpha}}(x_u(\alpha))\, dx_u(\alpha) - 1 \right) \\ &= - \ln q_{\boldsymbol{\psi}_{u\alpha}}(x_u(\alpha)) \\ &\quad + \mathbb{E}_{Q_{/\{x_u(\alpha)\}}} \sum_k \ln \mathcal{N}(\mathbf{f}_k(\mathbf{X}, \boldsymbol{\theta}) \mid \mathbf{m}_k, \boldsymbol{\Lambda}_k^{-1}) \\ &\quad + \mathbb{E}_{Q_{/\{x_u(\alpha)\}}} \ln \mathcal{N}(\mathbf{x}_u \mid \boldsymbol{\mu}_u, \boldsymbol{\Sigma}_u) + \lambda \end{aligned} \tag{13}$$

where the expectation is w.r.t. the proxy $Q_{/\{x_u(\alpha)\}} := \prod_{k \neq u} \prod_{t \neq \alpha} q_{\boldsymbol{\psi}_{kt}}(x_k(t))$, which we abbreviated with $\langle \cdot \rangle$ in the following. Bringing $\ln q_{\boldsymbol{\psi}_{u\alpha}}(x_u(\alpha))$ to the left side and using that the proxy distribution in equation 13 decomposes into a term that depends on the observations $\mathbf{Y}$ and a term that depends on the ODE's $\mathbf{f}(\mathbf{X}, \boldsymbol{\theta})$ we see that:

$$\begin{aligned} &\ln q_{\boldsymbol{\psi}_{u\alpha}}(x_u(\alpha)) := \ln q_{\boldsymbol{\psi}_{u\alpha}}(x_u(\alpha) \mid \boldsymbol{\theta}, \mathbf{Y}) \\ &\propto \underbrace{\langle \ln \mathcal{N}(\mathbf{x}_u \mid \boldsymbol{\mu}_u, \boldsymbol{\Sigma}_u) \rangle}_{=:\ln q(x_u(\alpha) \mid \mathbf{Y})} \\ &\underbrace{+ \sum_k \left\langle \ln \mathcal{N}\left( \mathbf{f}_k(\mathbf{X}, \boldsymbol{\theta}) \mid \mathbf{m}_k(\mathbf{X}), {\boldsymbol{\Lambda}^{(k)}}^{-1} \right) \right\rangle}_{=:\ln q(x_u(\alpha) \mid \boldsymbol{\theta})}. \end{aligned} \tag{14}$$

**Proposition 1** *$x_u(\alpha)$ appears quadratically in the proxy distribution that depends on $\mathbf{Y}$ in equation 14 and is thus Gaussian distributed:*

$$q(x_u(\alpha) \mid \mathbf{Y}) = \mathcal{N}(x_u(\alpha) \mid \iota, \Xi)$$

2. $Q(\mathbf{X})$ is a proxy to the posterior distribution $p(\mathbf{X} \mid \mathbf{Y}, \boldsymbol{\theta}, \boldsymbol{\phi}, \boldsymbol{\gamma})$.

**Proof** The proxy for $x_u(\alpha)$ given the observation $\mathbf{Y}$ is obtained by determining the conditional distribution of $x_u(\alpha)$ given all other time points of the same state and substituting their moments (Bishop, 2006, chapter 10)

$$q(x_u(\alpha) \mid \mathbf{Y}) = \mathcal{N}\left(x_u(\alpha) \mid \iota, \Xi\right), \tag{15}$$

with mean

$$\iota := \mu_u(\alpha) - \mathbf{\Sigma}^{(u)}_{\alpha\overline{\alpha}} \left(\mathbf{\Sigma}^{(u)}_{\overline{\alpha}\overline{\alpha}}\right)^{-1} \langle \mathbf{x}_{\overline{\alpha}} \rangle - \boldsymbol{\mu}_{\overline{\alpha}}$$

and variance

$$\Xi := \mathbf{\Sigma}_{\alpha\alpha} - \mathbf{\Sigma}_{\alpha\overline{\alpha}} \mathbf{\Sigma}^{-1}_{\overline{\alpha}\overline{\alpha}} \mathbf{\Sigma}_{\overline{\alpha}\alpha}.$$

The precise definition of $\mathbf{\Sigma}_{ij}$ can be found in definition 9 in the supplementary material 7. ■

Due to the specific functional form of the ODE 1, the states appear in quadratic in $\ln q(x_u(\alpha) \mid \boldsymbol{\theta})$, the second term in 14 and thus the proxy distribution is Gaussian distributed (Bishop, 2006, chapter 10.1.2):

**Proposition 2** *$x_u(\alpha)$ also appears quadratically in the proxy distribution that depends on $\boldsymbol{\theta}$ in equation 14 and is thus Gaussian distributed:*

$$q(x_u(\alpha) \mid \boldsymbol{\theta}) = \prod_k \mathcal{N}(x_u(\alpha) \mid \kappa_k, \Omega_k^2)$$

The proof with the precise form and definition of $\kappa_k$ and $\Omega_k^2$ is provided in the supplementary material 7.

**Lemma 3 (Proxy distribution)** *The mean $\nu_{u\alpha}(\mathbf{Y}, \boldsymbol{\theta})$ and variance $\Gamma_{u\alpha}(\mathbf{Y}, \boldsymbol{\theta})$, of the Gaussian proxy distribution $\widehat{q}(x_u(\alpha))$ are given by*

$$\begin{aligned} \Gamma_{u\alpha}(\mathbf{Y}, \boldsymbol{\theta}) &: = \left(\Xi^{-1} + \sum_k \left(\Omega_k^2\right)^{-1}\right)^{-1} \\ \nu_{u\alpha}(\mathbf{Y}, \boldsymbol{\theta}) &: = \Gamma_{u\alpha}(\mathbf{Y}, \boldsymbol{\theta}) \cdot \left(\Xi^{-1}\iota + \sum_k \left(\Omega_k^2\right)^{-1} \kappa_k\right). \end{aligned}$$

**Proof** Using proposition 1, proposition 2 and substituting both proxies into 14 with an according normalization yields:

$$\begin{aligned} q_{\psi_{u\alpha}}(x_u(\alpha)) &\propto q(x_u(\alpha) \mid \mathbf{Y}) \cdot q(x_u(\alpha) \mid \boldsymbol{\theta}) \\ &= \mathcal{N}(x_u(\alpha) \mid \iota, \Xi) \cdot \prod_k \mathcal{N}(x_u(\alpha) \mid \kappa_k, \Omega_k^2) \\ &\propto \mathcal{N}\left(x_u(\alpha) \mid \nu_{u\alpha}(\mathbf{Y}, \boldsymbol{\theta}), \Gamma_{u\alpha}(\mathbf{Y}, \boldsymbol{\theta})\right), \end{aligned}$$

The precise form of $\Gamma_{u\alpha}(\mathbf{Y}, \boldsymbol{\theta})$ and $\nu_{u\alpha}(\mathbf{Y}, \boldsymbol{\theta})$ as in 3 follows from the formula for the product of two Gaussian distributions (Petersen et al., 2008, section 8.1.8). ■

Therefore, one advantage over sampling methods is the determination of a proxy to the intractable posterior distribution over state variables given the observations and the ODE's:

**Lemma 4** *The proxy distribution over state variables given the observations and the ODE's is:*

$$p(\mathbf{X} \mid \mathbf{Y}, \boldsymbol{\theta}, \boldsymbol{\phi}, \boldsymbol{\gamma}) \approx \widehat{Q}(\mathbf{X}) = \prod_k \prod_t \widehat{q}_{\psi_{kt}}(x_k(t))$$

*with* $\widehat{q}_{\psi_{kt}}(x_k(t))$ *specified as above.*

In addition to the posterior in lemma 4, we have a proxy distribution for the logarithmic transform of the posterior over the parameters given the observations. Moreover, by the design of the proxy distribution $Q$, which decouples the state variables, we can in contrast to the integral over states in equation 9, analytically maximize the tractable lower bound $\mathcal{L}_Q(\boldsymbol{\theta})$ w.r.t. the ODE parameters $\boldsymbol{\theta}$ and summarize these results in lemma 5 and lemma 6:

**Lemma 5** *A log transformed proxy distribution over the parameters given the observations is:*

$$\begin{aligned} \ln p(\boldsymbol{\theta} \mid \mathbf{Y}, \boldsymbol{\phi}, \boldsymbol{\gamma}) &= \ln \int p(\boldsymbol{\theta} \mid \mathbf{X}, \boldsymbol{\theta}, \boldsymbol{\phi}, \boldsymbol{\gamma}) p(\mathbf{X} \mid \mathbf{Y}, \boldsymbol{\phi}) d\mathbf{X} \\ &\approx \mathcal{L}_{\widehat{Q}}(\boldsymbol{\theta}), \end{aligned} \tag{16}$$

*which was specified as the variational lower bound in equation 10, is analytically tractable.*

**Lemma 6** $\widehat{\boldsymbol{\theta}}$ *is Gaussian distributed:*

$$\widehat{\boldsymbol{\theta}} \sim \mathcal{N}(\widehat{\boldsymbol{\theta}} \mid \boldsymbol{\zeta}, \boldsymbol{\Psi}) \tag{17}$$

**Proof** Due to the functional form of the ODE in 2, $\mathbf{f}_k$ is linear in the parameters, which thus appear quadratic in $\sum_k \ln \mathcal{N}(\mathbf{f}_k(\mathbf{X}, \boldsymbol{\theta}) \mid \mathbf{m}_k, \boldsymbol{\Lambda}_k^{-1})$. Therefore the distribution is Gaussian and the precise formulas for the mean $\boldsymbol{\zeta}$ and variance $\boldsymbol{\Psi}$ follow as in the proof of proposition 2. Given the Gaussian distribution the maximum a posteriori estimator in equation 9 of $\boldsymbol{\theta}$ is achieved by setting it equal to the mean $\boldsymbol{\zeta}$. ■

We use an EM-like scheme to iterate between establishing tight variational lower bounds as described in lemma 5 and maximizing those lower bounds. The precise iterations between expectation and maximization steps until convergence are outlined in algorithm 1:

The "hill climbing" mechanism of mean-field variational inference is shown in figure 1.

**Algorithm 1** Mean-field variational inference for GP Gradient Matching

Initialization of maximum number of iteration $I_{\text{max}}$ for iteration index $i$
**repeat** $\forall \alpha \leq N$ and $\forall u \leq K$
Expectation step (E-step):
    Calculate $\nu_{u\alpha}(\mathbf{Y}, \hat{\boldsymbol{\theta}}^{(i)})$, $\Gamma_{u\alpha}(\mathbf{Y}, \hat{\boldsymbol{\theta}}^{(i)})$
Maximisation step (M-step):
    $\hat{\boldsymbol{\theta}}^{(i+1)} := \arg\max_{\boldsymbol{\theta}} \mathbb{E}_{\hat{Q}} \sum_k \ln \mathcal{N}(\mathbf{f}_k(\mathbf{X}, \boldsymbol{\theta}) \mid \mathbf{m}_k, \boldsymbol{\Lambda}_k^{-1})$
**until** Convergence = true or $i = I_{\text{max}}$

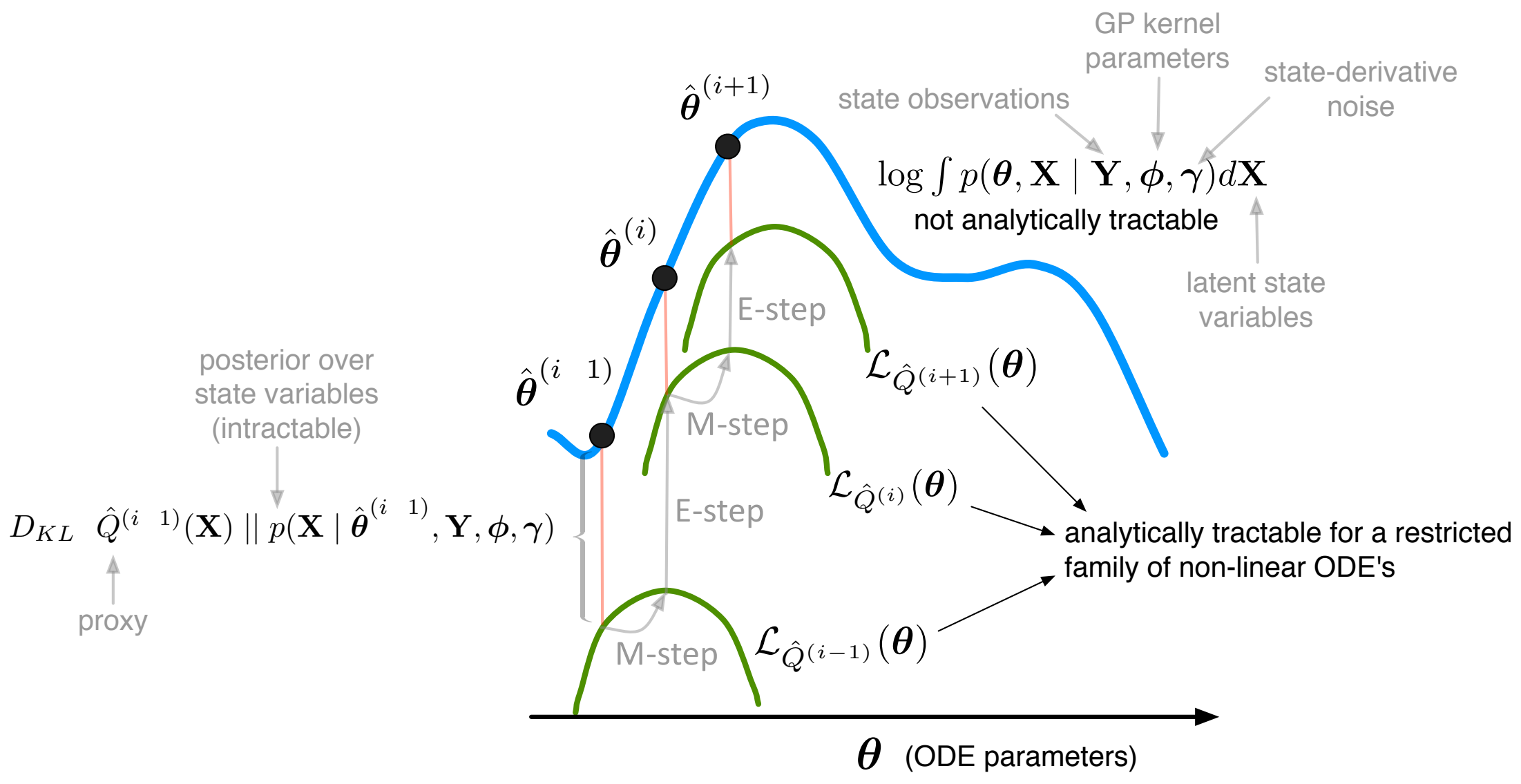

Figure 1: For a restricted family of ODE's mean-field variational inference establishes tight variational lower bounds $\mathcal{L}_{\widehat{Q}^{(\cdot)}}(\boldsymbol{\theta})$ that are analytically tractable and therefore facilitate "hill climbing" on the intractable log integral w.r.t. the ODE parameters $\boldsymbol{\theta}$. The difference between the lower bound $\mathcal{L}_{\widehat{Q}^{(\cdot)}}(\boldsymbol{\theta})$ and the log integral (red line) is given by the Kullback-Leibler divergence between the proxy distribution $\widehat{Q}^{(\cdot)}(\mathbf{X})$ and the intractable posterior distribution $p(\mathbf{X} \mid \mathbf{Y}, \boldsymbol{\theta}, \boldsymbol{\phi}, \boldsymbol{\gamma})$ as shown by the log integral identity 12.

**Remark 7**

- *In the E-step a tight lower bound is established by determining the moments,* $\nu_{u\alpha}(\mathbf{Y}, \widehat{\boldsymbol{\theta}}^{(i)})$ *and* $\Gamma_{u\alpha}(\mathbf{Y}, \widehat{\boldsymbol{\theta}}^{(i)})$ *of the proxy distribution* $\widehat{Q}^{(i)}$*, which are given in lemma 3.*

- *In the M-step, the analytically tractable lower bounds* $\mathcal{L}_{\widehat{Q}_i}$ *given in lemma 5 are maximized and we can not only provide the analytic solution* $\boldsymbol{\zeta}$ *for* $\widehat{\boldsymbol{\theta}}$ *but also its variance* $\boldsymbol{\Psi}$ *by lemma 6. In practice we did not explicitly calculate the mean* $\boldsymbol{\zeta}$ *but used gradient ascent, due to the concavity of the optimization problem.*

## 6. Experiments

In the following we test our approach on two small to medium sized ODE models. Both systems have been extensively studied, especially for gradient matching methods (e.g. Dondelinger et al., 2013; Wang and Barber, 2014). To simplify the comparison to other approaches, we try to use the same parameter settings whenever possible.

### 6.1 Lotka-Volterra System

The ODE $\mathbf{f}(\mathbf{X}, \boldsymbol{\theta})$ of the Lotka-Volterra system (Lotka, 1978) is given by:

$$\dot{x}_1 := \theta_1 x_1 - \theta_2 x_1 x_2$$
$$\dot{x}_2 := -\theta_3 x_2 + \theta_4 x_1 x_2$$

The above system is used to study predator-prey interactions and exhibits periodicity and non-linearity at the same time. We used the same ODE parameters as in Dondelinger et al. (2013) (i.e. $\theta_1 = 2, \theta_2 = 1, \theta_3 = 4, \theta_4 = 1$) to simulate the data over an interval $[0, 2]$ with a sampling interval of $0.1$. Predator species (i.e. $x_1$) were initialized to 3 and prey species (i.e. $x$) were initialized to 5. Mean field variational inference for gradient matching was performed on two such simulated datasets corrupted by additive standard Gaussian noise with variances $0.1$ and $0.25$. The radial basis function kernel was used to capture the covariance between a state at different time points. The trajectory of the proxy means (gray lines) w.r.t. the ODE parameters are shown in the top left and top right plots of figure 2 for the different Gaussian noise levels.

### 6.2 Protein Signalling Transduction Pathway

The chemical kinetics for the protein signalling transduction pathway is governed by a combination of mass action kinetics and the Michaelis-Menten kinetics and was first described in Vyshemirsky and Girolami (2008):

$$\begin{aligned}
[\dot{S}] &= -k_1 \times [S] - k_2 \times [S] \times [R] + k_3 \times [RS] \\
[\dot{S_d}] &= k_1 \times [S] \\
[\dot{R}] &= -k_2 \times [S] \times [R] + k_3 \times [RS] + V \times \frac{[Rpp]}{K_m + [Rpp]} \\
[\dot{RS}] &= k_2 \times [S] \times [R] - k_3 \times [RS] - k_4 \times [RS] \\
[\dot{Rpp}] &= k_4 \times [RS] - V \times \frac{[Rpp]}{K_m + [Rpp]}
\end{aligned} \tag{18}$$

We define the following latent variables:

$$x_1 := [S],\ x_2 := [S_d],\ x_3 := [R],\ x_4 := [RS],\ x_5 := \frac{[Rpp]}{K_m + [Rpp]}$$
$$\theta_1 := k_1, \theta_2 := k_2, \theta_3 := k_3, \theta_4 := k_4, \theta_5 := V$$

The transformation is motivated by the fact that in the new system, all states only appear as monomials. Thus our assumption in equation 2 is satisfied. However, the parameter $K_m$ is thus not identifiable and in addition we directly substitute $[\dot{Rpp}]$ with $[\dot{x_5}]$. The simplistic substitution is inconsistent e.g. with equation 5 and is used to show that our approach is even robust against model miss-specifications, as illustrated in figure 3.

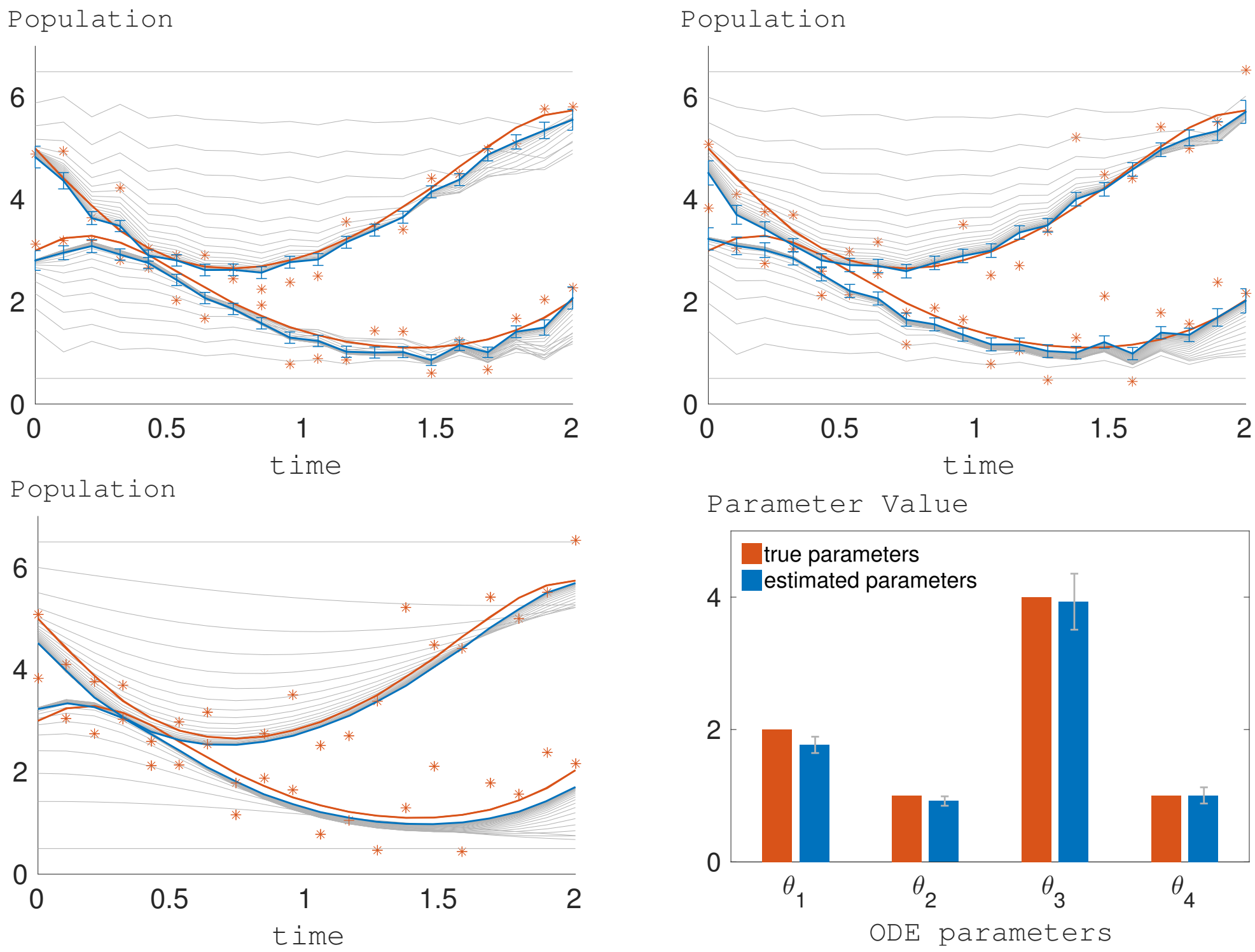

Figure 2: **Lotka-Volterra**: The top left and top right plots show the true state dynamics (red) together with the trajectory of the proxy means (gray lines) w.r.t. the ODE parameters for simulated data corrupted by standard Gaussian noise with variances $0.1$ and $0.25$, respectively. The blue lines represent the optimal proxy means. The bottom left plot shows the trajectory of the ODE solver with respect to the estimated ODE parameters (learned by mean-field variational inference) for the same data used in the top right plot. The bottom right plot shows that the estimated parameters (blue) learned by mean-field variational inference offer good approximations to the true parameters (red). Error bars represent one standard deviation of uncertainty.

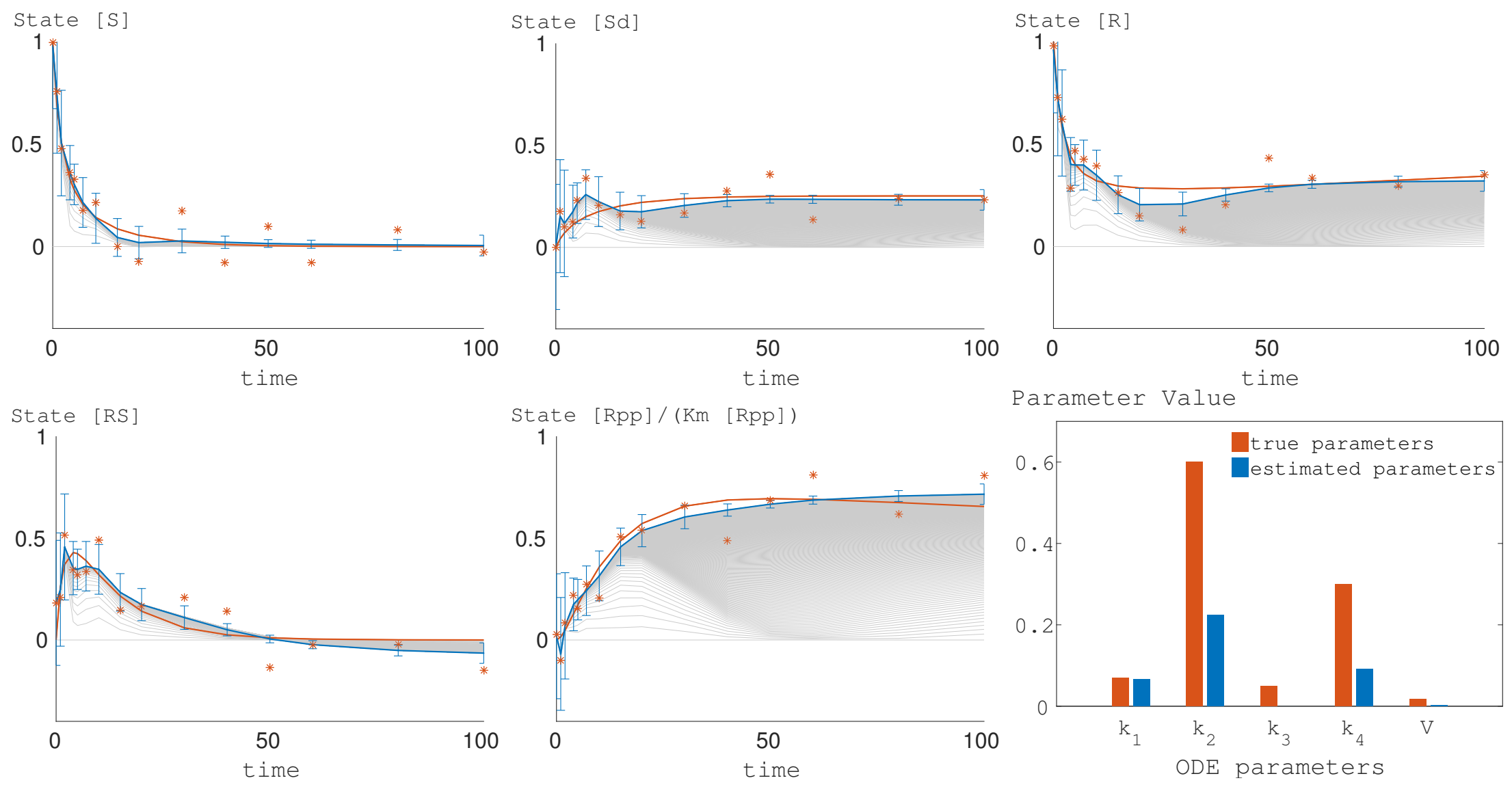

Figure 3: **Protein Transduction Pathway**: The quality of the parameter estimation is shown in the bottom right plot. While the parameters are not as well estimated as in the Lotka-Volterra model, the ranking of the parameter values is preserved. The remaining plots show that the mean of the proxy distribution (blue) estimates the true dynamics (red) of the states well. Error bars represent one standard deviation of uncertainty.

Once more, we use the same ODE parameters as in Dondelinger et al. (2013) i.e. $k_1 = 0.07, k_2 = 0.6, k_3 = 0.05, k_4 = 0.3, V = 0.017$. The data was sampled over an interval $[0, 100]$ with time point samples at $t = [0, 1, 2, 4, 5, 7, 10, 15, 20, 30, 40, 50, 60, 80, 100]$. Standard Gaussian noise with variance $0.01$ was used to corrupt the simulated data. The sigmoid kernel was used to capture the covariance between a state at different time points.

The parameters are not as well approximated as in example 6.1 for the Lotka-Volterra model. However the ranking according to parameter values is preserved as indicated by the bottom right plot in figure 3. In addition the remaining five plots in figure 3 show the trajectory of the proxy with respect to the ODE parameters. The mean of the proxy (blue) is able to capture the true dynamics of the states (red).

**Remark 8**

- *In our approach we initialize all proxy distributions and parameters by setting them to zero. Our approach works even when the initializations can be far away from the true value, as illustrated in figure 3 with the starting point of the converging gray lines. That an additional advantage, compared to previous work as in e.g. Dondelinger et al. (2013), where parameter values where initialized by Gaussian process regression.*

- *In addition, our approach is different to previous work because we provide a proxy to estimate the dynamics of the states, whereas previous work used an ODE solver to estimate the dynamics of the states. In this work, we provide a proxy to the dynamics of the state* without *using an ODE solver. Variational inference has been illustrated to work even for very high dimensional inference problems (e.g. Blei et al., 2016; Hoffman et al., 2013).*

## 7. Discussion

Numerical integration is a major bottleneck due to its computational cost for large scale parameter inference e.g. in systems biology. Whilst previous Gaussian process approaches circumvented the computational burden of numerical integration by providing parameters estimates without explicitly solving the ODE, our contribution of this paper is to integrate out the latent state variables instead of sampling them as in previous work. Since the integration over state variables is *not* analytically tractable we use mean-field approximation to establish tight variational lower bounds that decouple the state variables. Such tight variational lower bounds are analytically tractable provided the ODE is such that the state variables appear in quadratic form in equation 13. ODE's such as for example the Lotka-Volterra system full-fill these requirements. In an additional real-world example of signalling pathway identification, we show that our approach performs well, even when the variational bounds are not analytically tractable or under model miss-specifications. Moreover an additional advantage of variational inference over sampling approaches is the determination of a proxy $Q(\mathbf{X})$ for the intractable posterior distribution over state variables $\mathbf{X}$ given the state observations $\mathbf{Y}$ and the ODE's $\mathbf{f}(\mathbf{X}, \boldsymbol{\theta})$. Future extension of our approach cover the possible treatment of partially observable systems and approximations to systems where the lower bounds are not analytically tractable, yet. Moreover due to the speed and scalability of variational inference (e.g. Blei et al., 2016; Hoffman et al., 2013), we hope to evaluate our approach on large-scale systems. After estimating the parameters, previous work applied and ODE solver to estimate the dynamics of the states. In this work, we provide a proxy to the dynamics of the state *without* using an ODE solver.

### Acknowledgments

This research was partially supported by the Max Planck ETH Center for Learning Systems and the SystemsX.ch project SignalX.

# Supplement

**Definition 9**

$$\boldsymbol{\Sigma} =: \begin{pmatrix} \boldsymbol{\Sigma}_{\alpha\alpha} & \boldsymbol{\Sigma}_{\alpha\overline{\alpha}} \\ \boldsymbol{\Sigma}_{\overline{\alpha}\alpha} & \boldsymbol{\Sigma}_{\overline{\alpha}\overline{\alpha}} \end{pmatrix}, \text{of size} \begin{pmatrix} 1\times 1 & 1\times N-1 \\ (N-1)\times 1 & (N-1)\times(N-1) \end{pmatrix} \text{and}$$
$$\boldsymbol{\mu}_u =: \begin{pmatrix} \boldsymbol{\mu}_u(\alpha) \\ \boldsymbol{\mu}_u(\overline{\alpha}) \end{pmatrix}, \text{of size} \begin{pmatrix} 1 \\ N-1 \end{pmatrix},$$

*where $N$ is the number of observations.*

**Proof** [Proof Proposition 2] Our strategy in the following is to retain only terms in

$$\sum_k \left\langle \ln \mathcal{N}(\mathbf{f}_k(\mathbf{X},\boldsymbol{\theta},\mathbf{t}) \mid \mathbf{m}_k(\mathbf{X},\mathbf{t}), {\boldsymbol{\Lambda}^{(k)}}^{-1}) \right\rangle$$

that have a functional dependence on $x_u(\alpha)$[3].

$$\begin{aligned}
\mathbb{E}_{Q_{/\{x_u(\alpha)\}}} \sum_k \ln \mathcal{N}(\mathbf{f}_k(\mathbf{X},\boldsymbol{\theta}) \mid \mathbf{m}_k, \boldsymbol{\Lambda}_k^{-1}) &\propto \mathbb{E}_{Q_{/\{x_u(\alpha)\}}} \sum_k \mathcal{N}\left(\mathbf{F}_k\mathbf{x}_k \mid \mathbf{m}_k, \boldsymbol{\Lambda}_k^{-1}\right) \\
&= \mathbb{E}_{Q_{/\{x_u(\alpha)\}}} \sum_k \mathcal{N}\left(\mathbf{F}_k\mathbf{x}_k - \mathbf{m}_k \mid 0, \boldsymbol{\Lambda}_k^{-1}\right) \\
&= \sum_k \langle \mathcal{N}\left(\mathbf{B}_k\mathbf{x}_k \mid 0, \boldsymbol{\Lambda}_k^{-1}\right)\rangle
\end{aligned}$$

where $\mathbf{B}_k := (\mathbf{F}_k - \mathbf{m}_k) = \mathbf{F}_k -' \mathbf{C}_{\phi_k}\mathbf{C}_{\phi_k}^{-1}$. Here again we use $\langle\cdot\rangle$ to denote the expectation w.r.t. $Q_{/\{x_u(\alpha)\}} := \prod_{u\neq k}\prod_{t\neq\alpha} q_{\boldsymbol{\psi}_{kt}}(x(t))$. $\mathbf{F}_k$ is the matrix implicitly defined by $\mathbf{f}_k(\mathbf{X},\boldsymbol{\theta}) = \mathbf{F}_k\mathbf{x}_k + \mathbf{O}_k$, where $\mathbf{O}_k$ is a matrix without any dependence on $\mathbf{x}_k$. In the interest of legibility we drop the subscript of $k$ in the following.

$$\begin{aligned}
(\mathbf{B}\mathbf{x})^T\boldsymbol{\Lambda}\mathbf{B}\mathbf{x} = \mathbf{x}^T\mathbf{B}^T\boldsymbol{\Lambda}\mathbf{B}\mathbf{x} = \mathbf{x}^T\mathbf{B}^T\mathbf{B}\boldsymbol{\Lambda}\mathbf{x} &= \sum_t x_t \sum_{t'} \left(\mathbf{B}^T\mathbf{B}\boldsymbol{\Lambda}\right)_{tt'} x_{t'} = \sum_t\sum_{t'} x_t \left(\mathbf{B}^T\mathbf{B}\boldsymbol{\Lambda}\right)_{tt'} x_{t'} \\
&= \sum_t x_t^2 \left(\mathbf{B}^T\mathbf{B}\boldsymbol{\Lambda}\right)_{tt} + \sum_{t,t',t\neq t'} x_t \left(\mathbf{B}^T\mathbf{B}\boldsymbol{\Lambda}\right)_{tt'} x_{t'}
\end{aligned}$$

$$\begin{aligned}
\langle(\mathbf{B}\mathbf{x})^T\boldsymbol{\Lambda}\mathbf{B}\mathbf{x}\rangle &= \langle \sum_t x_t^2 \left(\mathbf{B}^T\mathbf{B}\boldsymbol{\Lambda}\right)_{tt} + \sum_{t,t',t\neq t'} x_t \left(\mathbf{B}^T\mathbf{B}\boldsymbol{\Lambda}\right)_{tt'} x_{t'}\rangle = \\
&= x_\alpha^2 \left(\mathbf{B}^T\mathbf{B}\boldsymbol{\Lambda}\right)_{\alpha\alpha} + \sum_{t\neq\alpha} \langle x_t^2 \left(\mathbf{B}^T\mathbf{B}\boldsymbol{\Lambda}\right)_{tt}\rangle + \sum_{t'\neq\alpha} x_\alpha \langle \left(\mathbf{B}^T\mathbf{B}\boldsymbol{\Lambda}\right)_{\alpha t'} x_{t'}\rangle \\
&+ \sum_{\substack{t,t'\in\mathbf{T}\setminus\{\alpha\} \\ t\neq t'}} \langle x_t \left(\mathbf{B}^T\mathbf{B}\boldsymbol{\Lambda}\right)_{tt'} x_{t'}\rangle \qquad (19)
\end{aligned}$$

3. Remaining terms only have an influence on the normalization.

Entries of $\mathbf{B}^T\mathbf{B}\mathbf{\Lambda}$ are given by

$$\begin{aligned}
\left(\mathbf{B}^T\mathbf{B}\mathbf{\Lambda}\right)_{ij} &= \left(\left(\mathbf{F}-\mathbf{E}\right)^T\left(\mathbf{F}-\mathbf{E}\right)\mathbf{\Lambda}\right)_{ij} = \left(\left(\mathbf{F}^T\mathbf{A} - 2\mathbf{F}^T\mathbf{E} + \mathbf{E}^T\mathbf{E}\right)\mathbf{\Lambda}\right)_{ij} \\
&= \left(\mathbf{F}^T\mathbf{F}\mathbf{\Lambda}\right)_{ij} - 2\left(\mathbf{F}^T\mathbf{E}\mathbf{\Lambda}\right)_{ij} + \left(\mathbf{E}^T\mathbf{E}\mathbf{\Lambda}\right)_{ij} \\
&= \sum_m (\mathbf{F}^T\mathbf{F})_{im}\mathbf{\Lambda}_{mj} - 2\sum_m \mathbf{F}^T_{im}\left(\mathbf{E}\mathbf{\Lambda}\right)_{mj} + \sum_m (\mathbf{E}^T\mathbf{E})_{im}\mathbf{\Lambda}_{mj} \\
&= \sum_m \mathbf{\Lambda}_{mj}\sum_l \mathbf{F}^T_{il}\mathbf{F}_{lm} - 2\sum_m \mathbf{F}^T_{im}\left(\mathbf{E}\mathbf{\Lambda}\right)_{mj} + \sum_m (\mathbf{E}^T\mathbf{E})_{im}\mathbf{\Lambda}_{mj}.
\end{aligned}$$

Substituting the corresponding values into equation 19 leads to

$$\langle \mathbf{x}^T\mathbf{B}^T\mathbf{B}\mathbf{\Lambda}\mathbf{x}\rangle = x_\alpha^2\langle \mathbf{W}_{\alpha\alpha}\rangle + x_\alpha\sum_{t'\neq\alpha}\langle \mathbf{W}_{\alpha t'}x_{t'}\rangle + \sum_{t\neq\alpha}\langle x_t^2\mathbf{W}_{tt}\rangle + \sum_{\substack{t,t'\in\mathbf{T}\setminus\{\alpha\}\\ t\neq t'}}\langle x_t\mathbf{W}_{tt'}x_{t'}\rangle, \quad (20)$$

with $\mathbf{W}_{ij} := (\mathbf{B}^T\mathbf{B}\mathbf{\Lambda})_{ij}$ i.e:

$$\langle \mathbf{W}_{\alpha\alpha}\rangle = \sum_m \mathbf{\Lambda}_{m\alpha}\sum_l\langle \mathbf{F}^T_{\alpha l}\mathbf{F}_{lm}\rangle - 2\sum_m\langle \mathbf{F}^T_{\alpha m}\rangle\left(\mathbf{E}\mathbf{\Lambda}\right)_{m\alpha} + \sum_m(\mathbf{E}^T\mathbf{E})_{\alpha m}\mathbf{\Lambda}_{m\alpha}$$

and

$$\langle \mathbf{W}_{\alpha t'}\mathbf{x}_{t'}\rangle = \sum_m \mathbf{\Lambda}_{mt'}\sum_l\langle \mathbf{F}^T_{\alpha l}\mathbf{F}_{lm}\mathbf{x}_{t'}\rangle - 2\sum_m\langle \mathbf{F}^T_{\alpha m}\mathbf{x}_{t'}\rangle\left(\mathbf{E}\mathbf{\Lambda}\right)_{mt'} + \sum_m(\mathbf{E}^T\mathbf{E})_{\alpha m}\mathbf{\Lambda}_{mt'}\langle \mathbf{x}_{t'}\rangle.$$

Due to the specific functional form of the ODE 1, equation 20 is a second order polynomial w.r.t. $x_u(\alpha)$ which implies that $q(x_u(\alpha) \mid \boldsymbol{\theta})$ is Gaussian distributed (Bishop, 2006, chapter 10.1.2). Thus we can determine the proxy distribution analytically:

$$q(x_u(\alpha) \mid \boldsymbol{\theta}) = \prod_k \mathcal{N}\left(x_u(\alpha) \mid \kappa_k, \Omega_k^2\right), \quad (21)$$

where the mean and variance are given by:

$$\Omega_k^2 := -\frac{1}{2}\langle \mathbf{W}_{\alpha\alpha}\rangle^{-1}$$

and

$$\begin{aligned}
\kappa_k := \Omega_k^2\sum_{t',t'\neq\alpha}\langle \mathbf{W}_{\alpha t'}\mathbf{x}_{t'}\rangle = \Omega_k^2\sum_{t',t'\neq\alpha}\sum_m \mathbf{\Lambda}_{mt'}\sum_l\langle \mathbf{F}^T_{\alpha l}\mathbf{F}_{lm}\mathbf{x}_{t'}\rangle - 2\sum_m\langle \mathbf{F}^T_{\alpha m}\mathbf{x}_{t'}\rangle\left(\mathbf{E}\mathbf{\Lambda}\right)_{mt'} \\
+ \sum_m(\mathbf{E}^T\mathbf{E})_{\alpha m}\mathbf{\Lambda}_{mt'}\langle \mathbf{x}_{t'}\rangle
\end{aligned}$$

■